\documentclass[conference]{IEEEtran}
\IEEEoverridecommandlockouts
\usepackage{cite}
\usepackage{amsmath,amssymb,amsfonts}
\usepackage{algorithmic}
\usepackage{graphicx}
\usepackage{textcomp}

\usepackage{booktabs}
\usepackage[table,xcdraw]{xcolor}
\usepackage{colortbl}
\usepackage{lipsum} 

\usepackage{xcolor}
\usepackage{tikz}
\usepackage{url}

\makeatletter
\newcommand{\linebreakand}{%
\end{@IEEEauthorhalign}
\hfill\mbox{}\par
\mbox{}\hfill\begin{@IEEEauthorhalign}
}
\makeatother

\def\BibTeX{{\rm B\kern-.05em{\sc i\kern-.025em b}\kern-.08em
    T\kern-.1667em\lower.7ex\hbox{E}\kern-.125emX}}
\begin{document}

\title{HyCARD-Net: A Synergistic Hybrid Intelligence Framework for Cardiovascular Disease Diagnosis\\

}
\author{
\IEEEauthorblockN{Rajan Das Gupta}
\IEEEauthorblockA{\textit{Faculty of Information Science and Technology} \\
\textit{American International University–Bangladesh} \\
Dhaka, Bangladesh \\
18-36304-1@student.aiub.edu}
\and
\IEEEauthorblockN{Xiaobin Wu}
\IEEEauthorblockA{\textit{School of Automotive Engineering} \\
\textit{Chengdu Industry and Trade College} \\
Chengdu, China \\
wuxiaobin@cdgmxy.edu.cn}
\linebreakand
\IEEEauthorblockN{Xun Liu\textsuperscript{*}} 
\IEEEauthorblockA{\textit{Faculty of Education} \\
\textit{Shinawatra University} \\
Bangkok, Thailand\\
\textsuperscript{*}Corresponding author: lxhkj2011@gmail.com}
\and
\IEEEauthorblockN{Jiaqi He}
\IEEEauthorblockA{\textit{Tilburg School of Social and Behavioral Sciences} \\
\textit{Tilburg University} \\
Tilburg, The Netherlands \\
hjqhejiaqi@gmail.com}
}

\maketitle

\begin{abstract}
Cardiovascular disease (CVD) remains the foremost cause of mortality worldwide, underscoring the urgent need for intelligent and data-driven diagnostic tools. Traditional predictive models often struggle to generalize across heterogeneous datasets and complex physiological patterns. To address this, we propose a hybrid ensemble framework that integrates deep learning architectures—Convolutional Neural Networks (CNN) and Long Short-Term Memory (LSTM)—with classical machine learning algorithms, including K-Nearest Neighbor (KNN) and Extreme Gradient Boosting (XGB), using an ensemble voting mechanism. This approach combines the representational power of deep networks with the interpretability and efficiency of traditional models. Experiments on two publicly available Kaggle datasets demonstrate that the proposed model achieves superior performance, reaching \textbf{82.30\%} accuracy on Dataset~I and \textbf{97.10\%} on Dataset~II, with consistent gains in precision, recall, and F$_1$-score. These findings underscore the robustness and clinical potential of hybrid AI frameworks for predicting cardiovascular disease and facilitating early intervention. Furthermore, this study directly supports the United Nations Sustainable Development Goal 3 (Good Health and Well-being) by promoting early diagnosis, prevention, and management of non-communicable diseases through innovative, data-driven healthcare solutions.
\end{abstract}

\begin{IEEEkeywords}
Cardiovascular disease prediction, heart disease diagnosis, hybrid machine learning, deep learning, convolutional neural network (CNN), long short-term memory (LSTM), ensemble learning, Sustainable Development Goal 3 (SDG 3): Good Health and Well-being, health informatics, AI for healthcare
\end{IEEEkeywords}

\section{Introduction}\label{sec:introduction}

The rapid growth of healthcare data has created both opportunities and challenges for accurate and timely clinical interpretation~\cite{bhatt2023effective, ramesh2022predictive}. The convergence of artificial intelligence (AI) and healthcare analytics has enabled advanced computational methods capable of revealing complex, nonlinear relationships in medical datasets. Among these, \textit{machine learning} (ML) algorithms have shown significant promise for analyzing patient data, identifying hidden correlations, and supporting early diagnosis and personalized treatment~\cite{nagavelli2022machine, aljammali2023prediction}.

According to the \textit{World Health Organization}, cardiovascular diseases (CVDs) remain the leading global cause of death, accounting for approximately 17.9 million deaths annually and representing nearly 31\% of global mortality~\cite{tsao2022heart, samavat2012programs}. Integrating AI-driven diagnostic models into healthcare systems has the potential to improve diagnostic accuracy, support early intervention, and optimize resource allocation~\cite{ogunpola2024machine, latha2019improving}. Consequently, developing frameworks that combine the representational power of deep learning with the interpretability of classical ML algorithms remains crucial for explainable and reliable CVD prediction.

In alignment with the \textit{United Nations Sustainable Development Goal~3 (Good Health and Well-being)}, which seeks to reduce premature mortality from non-communicable diseases by one-third by~2030, this study leverages artificial intelligence to advance the early diagnosis and prognosis of cardiovascular disease. By integrating explainable AI techniques with clinical data analytics, the proposed framework supports proactive detection and intervention strategies, thereby contributing to improved healthcare accessibility, equity, and sustainability.

Traditional ML algorithms such as Decision Trees, Support Vector Machines, and ensemble methods have been successfully applied to CVD prediction~\cite{bhatt2023effective, bakar2023review}. However, these methods often depend on manual feature engineering and may fail to capture higher-order temporal or spatial relationships. In contrast, \textit{deep learning} (DL) architectures automatically learn complex hierarchical features from raw clinical data, enhancing scalability and predictive precision~\cite{bhavekar2024heart, subramani2023cardiovascular}. Specifically, CNNs and LSTMs can model both spatial and sequential dependencies, making them suitable for heterogeneous physiological data~\cite{dutta2020efficient, goswami2023electrocardiogram, mehmood2021prediction}. To complement these, KNN and XGB provide interpretable and efficient decision-making mechanisms—yielding a balance between transparency and accuracy.

Recent research highlights that hybrid ML--DL models can effectively combine automatic feature learning and explainability~\cite{bhavekar2022hybrid, nazari2024detection, naser2024review}. For example, Saranya~\textit{et~al.}~\cite{Saranya2025DenseNetABiLSTM} proposed a DenseNet--Attention-BiLSTM architecture for multiclass arrhythmia detection using photoplethysmogram (PPG) signals, demonstrating that coupling convolutional and recurrent learning substantially improves diagnostic precision and generalization.
Such integration enhances diagnostic accuracy, robustness, and trustworthiness—key prerequisites for clinical adoption.

\begin{figure}[!t]
\centering
\includegraphics[width=\columnwidth]{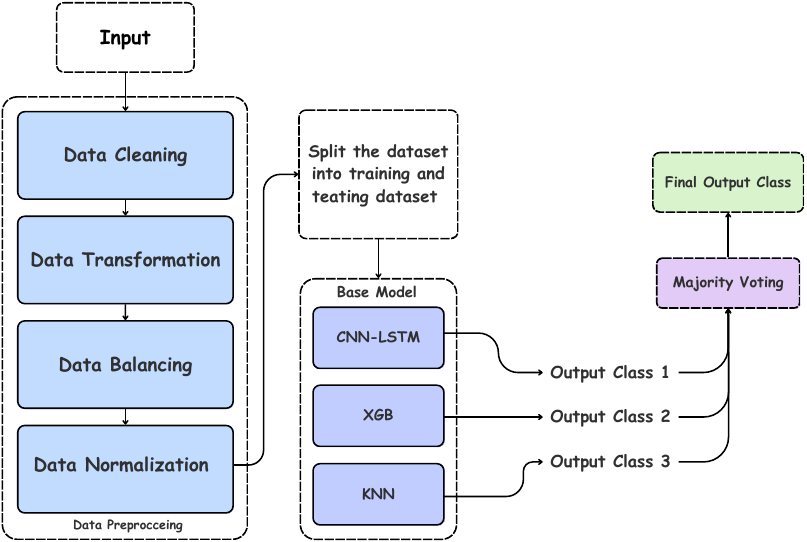}
\caption{Overall architecture of the proposed hybrid CNN--LSTM--KNN--XGB framework for cardiovascular disease prediction.}
\label{fig:cnn_lstm}
\end{figure}

This study proposes a \textit{hybrid diagnostic model} combining CNN and LSTM as the deep learning backbone with KNN and XGB as classical learners,as shown in Figure \ref{fig:cnn_lstm}. The final classification output is derived through an ensemble voting strategy that merges complementary model predictions for improved diagnostic consistency. Experiments on two public datasets demonstrate that the proposed model achieves superior predictive performance and strong generalization.

\textbf{Contributions:}
\begin{itemize}
    \item \textit{A hybrid CNN-LSTM + KNN + XGB architecture for high-accuracy CVD prediction.}
    \item \textit{Cross-dataset evaluation on two public and dataset demonstrating strong generalization.}
    \item \textit{A CVD-aware preprocessing pipeline with outlier filtering, AHA-based BP normalization, and class balancing.}
\end{itemize}

The remainder of this paper is organized as follows: Section~\ref{sec:related_work} reviews prior studies; Section~\ref{sec:methodology} presents the proposed model; Section~\ref{sec:results} discusses experimental results and comparative analysis; and Section~\ref{sec:conclusion} concludes the paper with future research directions.

\section{Related Work}\label{sec:related_work}

Machine learning (ML) and deep learning (DL) have become pivotal in cardiovascular disease (CVD) prediction owing to their ability to identify hidden risk factors, enhance diagnostic precision, and support data-driven decision-making~\cite{bhatt2023effective, bhavekar2024heart}. This section reviews representative studies across three categories—traditional ML, deep learning, and hybrid ML–DL approaches.

\subsection{Machine Learning Approaches}

Traditional ML algorithms have long been applied to structured medical datasets for CVD prediction. Ahmad~\textit{et al.}~\cite{ahamad2023influence} compared six classifiers—K-Nearest Neighbor (KNN), Logistic Regression (LR), Support Vector Machine (SVM), Random Forest (RF), Decision Tree (DT), and Extreme Gradient Boosting (XGB)—reporting SVM as most accurate (87.91\%) on the Cleveland dataset. Similarly, Akkaya~\textit{et al.}~\cite{akkaya2022comparative} evaluated eight ML methods and found KNN achieved 85.6\% accuracy. Feature-selection techniques have further improved classical models: Subanya and Rajalaxmi~\cite{subanya2014feature} optimized SVM via an Artificial Bee Colony algorithm (86.76\% accuracy), while Mokeddem~\textit{et al.}~\cite{mokeddem2013supervised} combined Genetic Algorithms with Naïve Bayes and SVM, achieving up to 85.5\%. Lin~\textit{et al.}~\cite{lin2023utilizing,lin2024tptm} used Taguchi-optimized ANN frameworks to enhance precision and reduce computational cost. Korial~\textit{et al.}~\cite{Korial2024improved} recently introduced an ensemble-based detection system with Chi-square feature selection, improving interpretability and feature robustness. Overall, ML techniques remain interpretable and efficient but rely heavily on manual feature engineering.

\subsection{Deep Learning Approaches}

Deep learning (DL) architectures automatically extract hierarchical features from raw clinical data, enabling improved representation of nonlinear and temporal patterns. Singhal~\textit{et al.}~\cite{singhal2018prediction} implemented a CNN with three convolutional layers on the Cleveland dataset and achieved 95\% accuracy, demonstrating the effect of network depth. Dutta~\textit{et al.}~\cite{dutta2020efficient} trained an efficient CNN on large-scale health survey data, reaching 81.78\% accuracy, while Mehmood~\textit{et al.}~\cite{mehmood2021prediction} proposed the \textit{CardioHelp} model integrating multiple CNN layers, achieving 97\% accuracy. Recent advances, such as Fan~\textit{et al.}~\cite{Fan2024cardio}, combined correlation analysis with ML to identify key biomarkers including troponin and CPK, while El~Massari~\textit{et al.}~\cite{Massari2024impact} incorporated ontology-driven feature representations to enhance CVD prediction accuracy. These studies confirm CNN’s ability to model spatial dependencies and LSTM’s strength in capturing sequential correlations—capabilities often beyond conventional ML models.

\subsection{Hybrid ML–DL Approaches}

Recent research increasingly integrates ML and DL to exploit their complementary advantages—automatic feature learning and interpretability. Tarawneh and Embarak~\cite{tarawneh2019hybrid} found that combining Naïve Bayes, SVM, DT, NN, and KNN in a hybrid model produced the best accuracy among individual classifiers. Bhavekar and Goswami~\cite{bhavekar2022hybrid} proposed an RNN–LSTM-based hybrid for cardiac classification, substantially improving precision. Subhadra and Vikas~\cite{subhadra2019neural} introduced a compact multilayer perceptron achieving 93.39\% accuracy. Building on this, Almutairi~\textit{et al.}~\cite{Almutairi2025intelligent} developed an intelligent hybrid modeling framework integrating ML and DL classifiers through a fusion mechanism, while Ganie~\textit{et al.}~\cite{Ganie2025ensemble} proposed an ensemble learning model coupled with SHAP-based explainability for transparent diagnosis. Ashika and Grace~\cite{Ashika2025enhancing} enhanced predictive accuracy using a stacked ensemble and multi-criteria decision-making (MCDM) approach with Rough Set Theory. Collectively, these hybrid strategies enhance both predictive power and robustness by merging data-driven representation with model interpretability.

\subsection{Summary and Research Gap}

While prior studies report high accuracies, most are dataset-specific and offer limited interpretability. Few address generalization across heterogeneous data or integrate deep and classical models in a balanced ensemble. Motivated by these gaps, this work introduces a unified hybrid framework combining CNN–LSTM (deep learners) and KNN–XGB (classical learners) within an ensemble voting scheme, designed to improve generalization, transparency, and reliability in CVD prediction.

\begin{figure}[!t]
\centering
\includegraphics[width=\columnwidth]{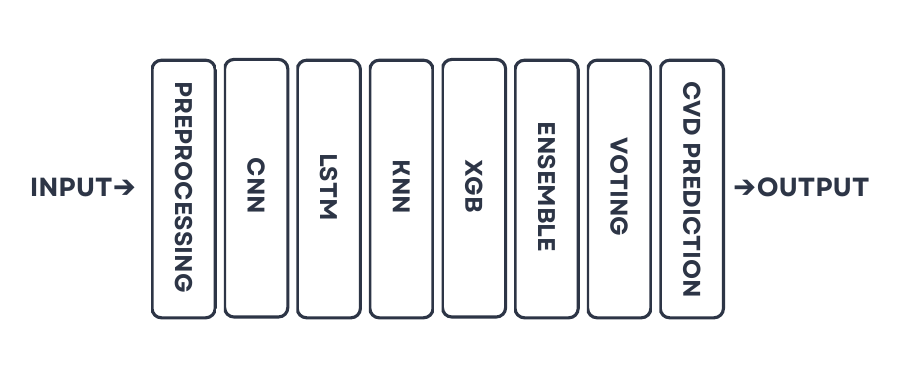}
\caption{Hybrid CNN–LSTM, KNN, and XGB Architecture for CVD Prediction}
\label{fig:cnn_cvd}
\end{figure}

\section{Proposed Model}\label{sec:methodology}

Machine learning (ML) and deep learning (DL) methods \ref{fig:cnn_lstm} offer complementary capabilities for cardiovascular disease (CVD) prediction. Classical ML models such as K-Nearest Neighbor (KNN) and Extreme Gradient Boosting (XGB) provide interpretability and computational efficiency, whereas deep architectures like Convolutional Neural Networks (CNN) and Long Short-Term Memory (LSTM) networks automatically learn complex nonlinear representations from raw data. Integrating these paradigms yields a balanced diagnostic framework that unites feature learning and explainable decision-making.

The proposed hybrid model combines two DL classifiers (CNN and LSTM) and two ML classifiers (KNN and XGB), as illustrated in Figure \ref{fig:cnn_cvd}. Each produces an independent prediction, and the final output is determined through ensemble voting weighted by each model’s validation accuracy. This strategy leverages model diversity to improve robustness and generalization across heterogeneous patient data.

\subsection{Model Rationale}
Each component contributes unique analytical strengths. CNN autonomously extracts discriminative spatial or structural features, while LSTM models temporal dependencies in sequential or longitudinal data. KNN, a non-parametric learner, identifies local similarity patterns, and XGB captures nonlinear feature interactions through gradient boosting. Together, these algorithms provide a comprehensive and interpretable diagnostic view.

\subsection{CVD-Specific Design Considerations}
The framework design reflects key characteristics of cardiovascular disease (CVD). Temporal variation in physiological indicators (e.g., blood pressure, glucose) motivated the use of LSTM for sequential modeling. Nonlinear and heterogeneous feature interactions required CNN for discriminative feature extraction. Local similarity patterns in clinical profiles supported the inclusion of KNN. Additionally, complex risk-factor interactions justified the use of XGB to capture nonlinear decision boundaries. These CVD-aware considerations guided the hybrid architecture.

\subsection{Data Preprocessing}
Before training, all datasets were standardized through the following steps:
\begin{itemize}
    \item \textbf{Data cleaning:} Duplicate and inconsistent records were removed, and missing values handled.
    \item \textbf{Transformation and normalization:} Categorical features were encoded and numerical values scaled to a uniform range.
    \item \textbf{Balancing:} Synthetic oversampling was applied to mitigate class imbalance and prevent bias toward the majority class.
\end{itemize}

Cleaned data were partitioned into 80\% training and 20\% testing subsets.

\subsection{Hybrid Architecture and Ensemble Strategy}
The CNN–LSTM combination forms the primary DL backbone, where CNN layers extract local representations and LSTM units capture sequential dependencies~\cite{taye2023understanding}. XGB and KNN act as complementary classical learners operating in parallel. Their predictions are integrated via weighted majority voting, ensuring that each model’s contribution reflects its validation performance.

Formally, the ensemble output is computed as:
\begin{equation}
\hat{y} = \arg\max_{c \in \mathcal{C}} \sum_{i=1}^{N} w_i\,\mathbb{I}\big(h_i(x) = c\big),
\end{equation}
where $h_i(x)$ is the prediction of the $i^{\text{th}}$ base classifier, $w_i$ its validation-derived weight, $\mathcal{C}$ the set of class labels, and $\mathbb{I}(\cdot)$ the indicator function.

\subsection{Computational Cost and Model Complexity}
All experiments were executed in Python~3.10 on Google Colab using GPU acceleration when available (Intel Core i7 CPU, 32 GB RAM). Table \ref{tab:cost} summarizes approximate parameter counts and average training times per fold. The hybrid ensemble adds minimal overhead compared with standalone DL models while yielding substantial performance gains.

\begin{table}[!t]
\caption{Runtime and model complexity per fold (Dataset II).}
\label{tab:cost}
\centering
\renewcommand{\arraystretch}{1.1}
\setlength{\tabcolsep}{5pt}
\begin{tabular}{lcc}
\toprule
\textbf{Model} & \textbf{Params (approx.)} & \textbf{Train time (s)} \\
\midrule
KNN & -- & 2.1 \\
XGB & $\sim$0.2 M & 8.4 \\
CNN & $\sim$0.6 M & 22.7 \\
LSTM & $\sim$0.5 M & 24.1 \\
CNN--LSTM & $\sim$1.1 M & 38.5 \\
\textbf{Hybrid (proposed)} & $\sim$\textbf{1.3 M} & \textbf{45.2} \\
\bottomrule
\end{tabular}
\end{table}

Overall, the hybrid ensemble unites the representation power of deep learning with the interpretability of traditional ML, forming a transparent and robust diagnostic system for CVD prediction.

\definecolor{headergray}{gray}{0.85}
\definecolor{rowgray}{gray}{0.94}

\begin{table*}[!t]
\centering
\caption{Evaluation metrics comparison on Dataset~I.}
\label{tab6}
\renewcommand{\arraystretch}{1.15}
\setlength{\tabcolsep}{6pt}
\begin{tabular*}{\textwidth}{@{\extracolsep{\fill}} l l cccc}
\rowcolor{headergray}
\textbf{Group} & \textbf{Model} & \textbf{Accuracy (std)} & \textbf{Recall (std)} & \textbf{F$_1$ (std)} & \textbf{Precision (std)} \\ 
\midrule
\rowcolor{rowgray}
\textbf{Machine learning} & NB & 75.50 (0.13) & 75.90 (0.08) & 76.00 (0.17) & 75.80 (0.11) \\
 & LR & 76.80 (0.08) & 77.80 (0.28) & 77.40 (0.21) & 77.20 (0.15) \\
\rowcolor{rowgray}
 & SVM & 77.50 (0.29) & 77.90 (0.15) & 77.70 (0.19) & 77.80 (0.08) \\
 & RF & 77.80 (0.11) & 78.10 (0.19) & 78.20 (0.23) & 78.30 (0.25) \\
\rowcolor{rowgray}
 & MLP & 76.80 (0.14) & 77.90 (0.11) & 77.80 (0.09) & 77.60 (0.23) \\
 & KNN & 78.30 (0.16) & 79.10 (0.17) & 79.20 (0.16) & 79.40 (0.12) \\
\rowcolor{rowgray}
 & DT & 77.30 (0.28) & 78.20 (0.12) & 78.30 (0.14) & 78.40 (0.05) \\
\textbf{Ensemble-based} & XGB & 78.90 (0.31) & 79.30 (0.23) & 79.20 (0.14) & 79.40 (0.22) \\
\rowcolor{rowgray}
 & LGBM & 78.10 (0.17) & 78.90 (0.28) & 78.60 (0.21) & 78.70 (0.17) \\
 & AdaBoost & 77.90 (0.23) & 78.80 (0.23) & 78.70 (0.19) & 78.90 (0.26) \\
\rowcolor{rowgray}
\textbf{Combinational ML} & RF + DT + NB + LR & 79.50 (0.19) & 79.70 (0.23) & 79.40 (0.09) & 79.50 (0.20) \\
 & RF + DT + KNN + NB + SVM + LGBM + XGB + MLP + LR & 79.00 (0.28) & 79.20 (0.20) & 79.00 (0.08) & 79.10 (0.11) \\
\rowcolor{rowgray}
\textbf{Deep learning} & CNN & 78.80 (0.32) & 80.00 (0.15) & 79.80 (0.13) & 80.10 (0.14) \\
 & LSTM & 78.00 (0.19) & 80.10 (0.27) & 79.90 (0.17) & 80.00 (0.27) \\
\rowcolor{rowgray}
\textbf{Combinational DL} & CNN--LSTM & 80.00 (0.21) & 81.80 (0.14) & 81.40 (0.08) & 81.30 (0.14) \\
\textbf{Proposed model} & CNN--LSTM + KNN + XGB & \textbf{82.30 (0.23)} & \textbf{84.00 (0.09)} & \textbf{83.50 (0.18)} & \textbf{83.60 (0.11)} \\
\bottomrule
\end{tabular*}
\end{table*}

\begin{table*}[!t]
\centering
\caption{Evaluation metrics comparison on Dataset~II.}
\label{tab7}
\renewcommand{\arraystretch}{1.15}
\setlength{\tabcolsep}{6pt}
\begin{tabular*}{\textwidth}{@{\extracolsep{\fill}} l l cccc}
\rowcolor{headergray}
\textbf{Group} & \textbf{Model} & \textbf{Accuracy (std)} & \textbf{Recall (std)} & \textbf{F$_1$ (std)} & \textbf{Precision (std)} \\ 
\midrule
\rowcolor{rowgray}
\textbf{Machine learning} & NB & 90.50 (0.11) & 90.20 (0.14) & 91.00 (0.28) & 91.40 (0.21) \\
 & LR & 91.70 (0.28) & 91.80 (0.13) & 91.60 (0.21) & 92.10 (0.17) \\
\rowcolor{rowgray}
 & SVM & 90.80 (0.19) & 91.50 (0.18) & 91.40 (0.24) & 91.70 (0.13) \\
 & RF & 91.90 (0.16) & 92.10 (0.24) & 92.30 (0.15) & 92.40 (0.15) \\
\rowcolor{rowgray}
 & MLP & 93.50 (0.14) & 93.60 (0.15) & 93.40 (0.19) & 93.70 (0.22) \\
 & KNN & 94.35 (0.13) & 94.50 (0.05) & 94.80 (0.19) & 94.60 (0.14) \\
\rowcolor{rowgray}
 & DT & 91.75 (0.08) & 92.60 (0.21) & 92.40 (0.21) & 92.70 (0.28) \\
\textbf{Ensemble-based} & XGB & 94.10 (0.09) & 94.00 (0.15) & 93.80 (0.12) & 94.00 (0.08) \\
\rowcolor{rowgray}
 & LGBM & 92.80 (0.21) & 93.70 (0.18) & 93.50 (0.31) & 93.20 (0.23) \\
 & AdaBoost & 91.50 (0.20) & 92.70 (0.28) & 92.40 (0.19) & 92.80 (0.24) \\
\rowcolor{rowgray}
\textbf{Combinational ML} & RF + DT + NB + LR & 94.50 (0.31) & 92.10 (0.08) & 93.10 (0.23) & 92.90 (0.19) \\
 & RF + DT + KNN + NB + SVM + LGBM + XGB + MLP + LR & 94.30 (0.05) & 93.50 (0.21) & 93.20 (0.48) & 93.40 (0.14) \\
\rowcolor{rowgray}
\textbf{Deep learning} & CNN & 93.80 (0.19) & 93.60 (0.13) & 93.70 (0.13) & 93.50 (0.09) \\
 & LSTM & 91.50 (0.16) & 92.60 (0.14) & 92.80 (0.15) & 92.70 (0.21) \\
\rowcolor{rowgray}
\textbf{Combinational DL} & CNN--LSTM & 94.45 (0.15) & 94.70 (0.13) & 94.90 (0.11) & 94.70 (0.12) \\
\textbf{Proposed model} & CNN--LSTM + KNN + XGB & \textbf{97.10 (0.17)} & \textbf{95.50 (0.28)} & \textbf{95.20 (0.08)} & \textbf{96.80 (0.14)} \\
\bottomrule
\end{tabular*}
\end{table*}

\section{Experiments and Results}\label{sec:results}

\subsection{Datasets}
Two publicly available heart-disease datasets from Kaggle were used for evaluation. Each underwent identical preprocessing to ensure consistency.

\textbf{Dataset I} (\url{https://www.kaggle.com/datasets/sulianova/cardiovascular-disease-dataset}) contains 70 000 clinical records with 12 attributes—11 input features and one binary output representing disease presence. After duplicate and extreme-outlier removal, 62 267 valid samples remained. The \textit{age} feature was converted from days to years, and systolic/diastolic pressures were categorized per American Heart Association guidelines.

\textbf{Dataset II} (\url{https://www.kaggle.com/datasets/mexwell/heart-disease-dataset}) merges five repositories: Cleveland (303), Hungarian (294), Switzerland (123), Long Beach VA (200), and Statlog (270). After deduplication and handling of missing values, 918 records across 12 attributes were retained.

\subsection{Evaluation Metrics}
Model performance was evaluated using five standard classification metrics: accuracy, precision, recall, F$_1$-score, and specificity.  
Each metric highlights a distinct aspect of predictive behavior.
\begin{equation}
\mathrm{Accuracy}=\frac{TP+TN}{TP+TN+FP+FN}
\end{equation}

Overall proportion of correctly classified samples.
\begin{equation}
\mathrm{Precision}=\frac{TP}{TP+FP}
\end{equation}

Correctly identified positives among all predicted positives.
\begin{equation}
\mathrm{Recall}=\frac{TP}{TP+FN}
\end{equation}

Sensitivity—ability to capture actual positive cases.
\begin{equation}
F_{1}=2\times\frac{\mathrm{Precision}\times\mathrm{Recall}}{\mathrm{Precision}+\mathrm{Recall}}
\end{equation}

Harmonic mean of precision and recall.
\begin{equation}
\mathrm{Specificity}=\frac{TN}{TN+FP}
\end{equation}

Capability to correctly identify negative instances.

Here, $TP$, $TN$, $FP$, and $FN$ denote true positives, true negatives, false positives, and false negatives.  
All metrics were computed per fold and averaged over ten cross-validation iterations.

\subsection{Implementation Details}
Each dataset was randomly partitioned into 80 \% training and 20 \% testing subsets. Ten-fold cross-validation was applied on the training set, with oversampling performed only within training folds to prevent data leakage. Performance metrics were averaged across folds. All hyperparameters were tuned using grid search to maximize validation accuracy. Unless otherwise stated, improvements over the strongest baseline were statistically significant under a paired $t$-test ($p<0.05$).

\subsubsection*{Cross-Validation and Standard Deviation Computation}
All results are reported as the mean and standard deviation (std) from 10-fold cross-validation on the training set. Oversampling was applied only within each training fold to prevent data leakage. Each fold was trained and evaluated independently, and the std values capture metric variability across the ten runs, ensuring reliable and unbiased performance estimation.

\subsection{Ablation Study}
As shown in Table~\ref{tab:ablation_dataset1} and Table~\ref{tab:ablation}, the CNN and LSTM baselines provide comparable starting performance, while the CNN-LSTM model yields a clear improvement on both datasets. The full hybrid configuration achieves the highest accuracy, confirming the cumulative benefit of integrating deep feature extraction with ensemble-based learning.

\subsection{Performance Evaluation}
The hybrid framework was compared with individual ML and DL baselines across both datasets.  
Average ten-fold cross-validation results with standard deviations are summarized in Tables \ref{tab6} and \ref{tab7}.  

\textbf{Key Findings:}
\begin{enumerate}
    \item Among traditional classifiers, KNN achieved the highest accuracy, while XGB performed best among ensemble-based methods.  
    \item Hybrid models consistently outperformed single algorithms, confirming the benefit of integrating multiple learning paradigms.  
    \item Deep-learning architectures, particularly the proposed hybrid ensemble, achieved the highest overall accuracy by effectively capturing complex nonlinear feature interactions.  
    \item The proposed model demonstrated the strongest discriminative performance and stable generalization across datasets.
\end{enumerate}

\begin{table}[!t]
\caption{Ablation on Dataset I (Accuracy \%).}
\label{tab:ablation_dataset1}
\centering
\renewcommand{\arraystretch}{1.1}
\setlength{\tabcolsep}{5pt}
\begin{tabular}{lc}
\toprule
\textbf{Configuration} & \textbf{Accuracy} \\
\midrule
CNN & 78.2 \\
LSTM & 78.0 \\
CNN--LSTM & 81.0 \\
\textbf{Hybrid (proposed)} & \textbf{82.3} \\
\bottomrule
\end{tabular}
\end{table}

\begin{table}[!t]
\caption{Ablation on Dataset II (Accuracy \%).}
\label{tab:ablation}
\centering
\renewcommand{\arraystretch}{1.1}
\setlength{\tabcolsep}{5pt}
\begin{tabular}{lc}
\toprule
\textbf{Configuration} & \textbf{Accuracy} \\
\midrule
CNN & 93.8 \\
LSTM & 91.5 \\
CNN--LSTM & 94.5 \\

\textbf{Hybrid (proposed)} & \textbf{97.1} \\
\bottomrule
\end{tabular}
\end{table}

\section{Discussion}\label{sec:discussion}

Accurate prediction of cardiovascular disease (CVD) risk is essential for early diagnosis and proactive treatment. The proposed hybrid ensemble effectively combines the representational depth of deep learning with the interpretability of classical machine learning models. Experimental results across two benchmark datasets confirmed that this integration yields consistent improvements in all major evaluation metrics—accuracy, recall, F$_1$-score, and precision. 

The superior performance of the hybrid model arises from its ability to capture complementary feature representations: CNN and LSTM extract spatial and temporal dependencies, while KNN and XGB provide localized and gradient-based decision boundaries. This combination produces a balanced decision-support framework that enhances both predictive reliability and explainability—two crucial aspects for clinical implementation.

Although results demonstrate strong generalization, real-world deployment requires further validation and compliance with clinical standards. Variations in population demographics, hospital procedures, and data acquisition can affect generalization performance. Future extensions may focus on model adaptability and interpretability through explainable AI (XAI) methods such as SHAP or LIME, enabling clinicians to better understand feature contributions during prediction.

\section{Conclusion}\label{sec:conclusion}

Cardiovascular disease remains a major global health burden, underscoring the need for accurate and interpretable predictive tools. This work introduced a hybrid framework that integrates CNN--LSTM feature extraction with KNN and XGB classifiers under an ensemble voting mechanism. The model achieved up to \textbf{97.10\%} accuracy across two public datasets, consistently outperforming all individual baselines. The findings demonstrate that combining deep-learning representations with interpretable machine-learning decision rules yields a robust and scalable solution for CVD prediction, supporting improved patient stratification and clinical decision-making.

This research also aligns with \textit{UN SDG~3 (Good Health and Well-being)}, contributing to the development of AI-driven diagnostic systems for early cardiovascular risk detection and broader healthcare accessibility. Future work will focus on evaluating the framework on larger real-world datasets, incorporating real-time physiological signals, and integrating explainable AI techniques to enhance transparency and clinical trust.

In summary, the proposed hybrid architecture provides a strong and interpretable foundation for next-generation AI-assisted cardiovascular diagnostics and supports the advancement of smart, data-driven healthcare environments.

\bibliographystyle{IEEEtran}
\bibliography{reference}

\end{document}